\documentclass[letterpaper, 10 pt, conference]{ieeeconf}  
\IEEEoverridecommandlockouts 
\usepackage{subfigure}
\usepackage{color}
\usepackage{graphicx}    
\usepackage{float}
\usepackage{multirow}
\usepackage{algorithm}
\usepackage{algpseudocode}
\usepackage{amsmath}
\usepackage{amssymb,amsfonts}
\usepackage[top=2cm, bottom=2cm, left=2cm, right=2cm]{geometry}
\usepackage{algorithmicx}
\usepackage{algpseudocode}
\usepackage{hyperref}
\usepackage{threeparttable}
\usepackage{booktabs}

\usepackage{indentfirst}
\usepackage{amsmath}
\usepackage{cite}

\usepackage{color}
\usepackage{soul}
\soulregister\cite7
\soulregister\ref7

\title{\LARGE \bf
ReactVLA: Fast and Lightweight Reactive Robot Manipulation via Improved Mean Flow Action Generation
}

\author{Yanzhao Guo$^{1}$, Wenkai Chen$^{2}$,  Jianwei Zhang$^{2}$%
\thanks{$^{1}$Shanghai Jiao Tong University}%
\thanks{$^{2}$Technical Aspects of Multimodal Systems (TAMS), Department of Informatics, Universit\"{a}t Hamburg. email: wenkai.chen@uni-hamburg.de}
}

\begin{document}
\maketitle
\thispagestyle{empty}
\pagestyle{empty}

\begin{abstract}
Diffusion-based Vision-Language-Action (VLA) policies have demonstrated strong capability in modeling expressive and multimodal action distributions. However, their reliance on iterative sampling introduces substantial inference latency, which limits their applicability to reactive closed-loop robot manipulation. To address this limitation, we propose \texttt{ReactVLA}, a lightweight and low-latency VLA framework for real-time robotic manipulation. \texttt{ReactVLA} combines two complementary designs: (1) an improved Mean Flow (iMF) action generator that reduces expensive multi-step diffusion sampling to one-to-few-step action generation, and (2) Attention Residuals (AttnRes), a dynamic depth-wise feature routing mechanism that replaces uniform residual accumulation to better preserve task-relevant multimodal representations. We evaluate \texttt{ReactVLA} on large-scale simulation benchmarks, including LIBERO and RoboIMI, as well as real-world robotic manipulation tasks. Experimental results show that \texttt{ReactVLA} consistently outperforms similarly sized VLA baselines, including SmolVLA and $\pi_0$. On challenging precision manipulation tasks, \texttt{ReactVLA} achieves up to a 1.65$\times$ improvement in task performance while providing more than a 4$\times$ increase in inference speed compared with leading VLA models. Finally, it reduces real-world policy latency to below 38.6 ms, enabling fast reactive control on physical robot platforms. Please check out our project website at: \url{https://game-loader.github.io/ReactVLA/}.
\end{abstract}

\section{Introduction}
\label{sec:introduction}

Learning-based robot manipulation increasingly relies on expressive generative policies to model the complex, multimodal action distributions arising from human demonstrations.
Diffusion-based visuomotor policies, in particular, have demonstrated remarkable manipulation performance by formulating action generation as an iterative denoising process~\cite{chi2023diffusion}, inheriting the strong representational capacity of modern diffusion generative models~\cite{ho2020denoising}.
Recent Vision-Language-Action (VLA) systems further scale this paradigm through large multimodal transformers, enabling instruction-conditioned robotic control across diverse manipulation tasks. Despite these advances, generative robot policies remain fundamentally constrained by inference latency.
Standard diffusion-style action generation requires dozens of sequential denoising evaluations during deployment, often resulting in inference times of hundreds of milliseconds per control cycle~\cite{black2024pi0}.
Such latency severely limits the high-frequency closed-loop interaction required for reactive robot manipulation, where policies must continuously adapt to contact dynamics, object motion, and execution uncertainty in real time. This raises a central question: can generative robot policies preserve the expressive advantages of diffusion-style action modeling while reducing inference to only one or a few evaluations?
In this work, we address this challenge through \texttt{ReactVLA}, a low-latency generative manipulation framework designed for reactive VLA control.

\begin{figure}[t]
  \centering
  \includegraphics[width=0.7\columnwidth]{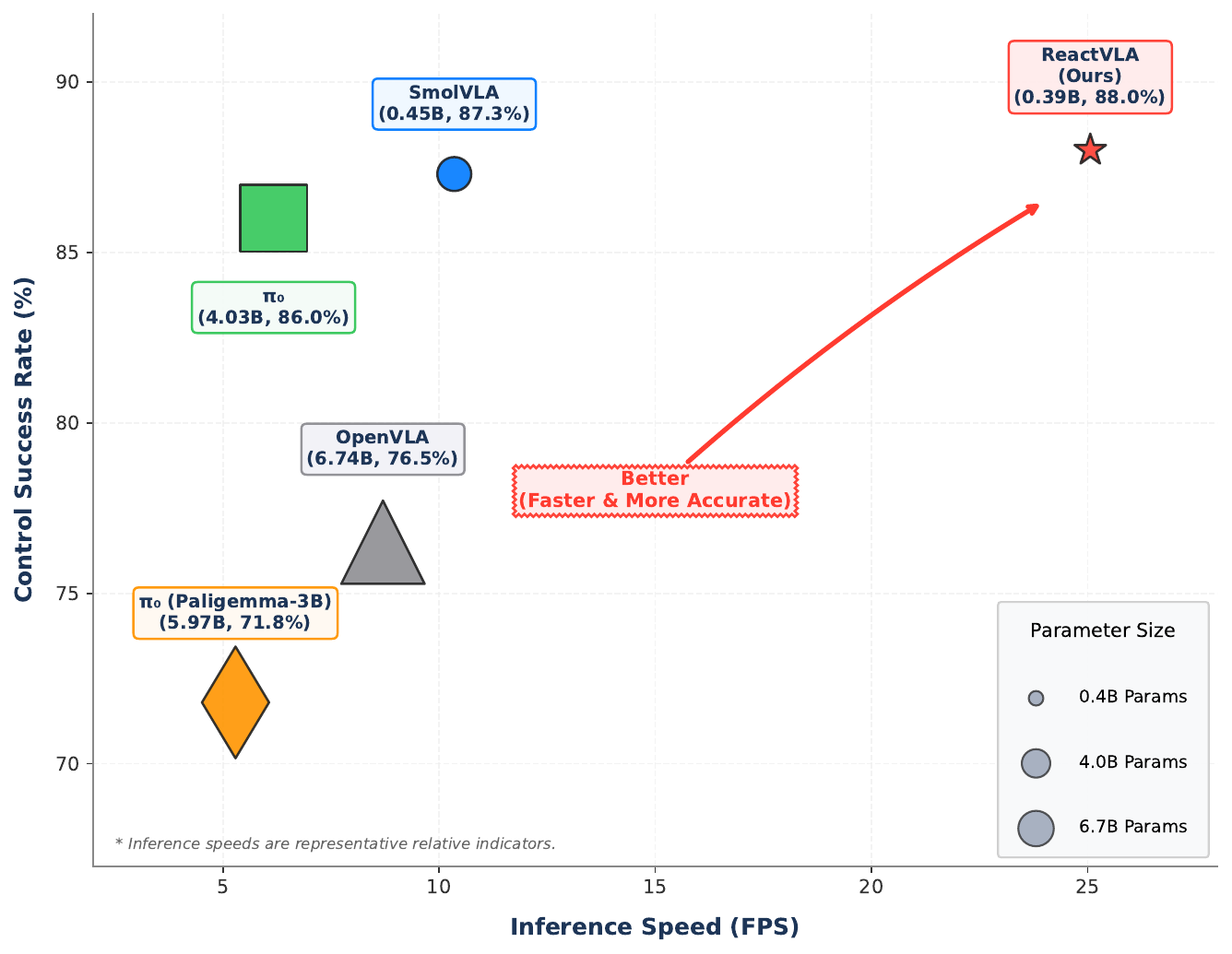}
  \caption{\textbf{Comparison of VLA Models across Model Size, Inference Latency, and Task Success Rate.} We compare \texttt{ReactVLA} against competitive VLA baselines (such as SmolVLA and $\pi_0$). Our method achieves the best average success rate among compared lightweight VLA models while maintaining a low inference latency, providing a practical trade-off for real-time reactive robotic control.}
  \label{fig:lightweight}
\end{figure}
\begin{figure*}[t]
    \centering
    \includegraphics[width=0.8\textwidth]{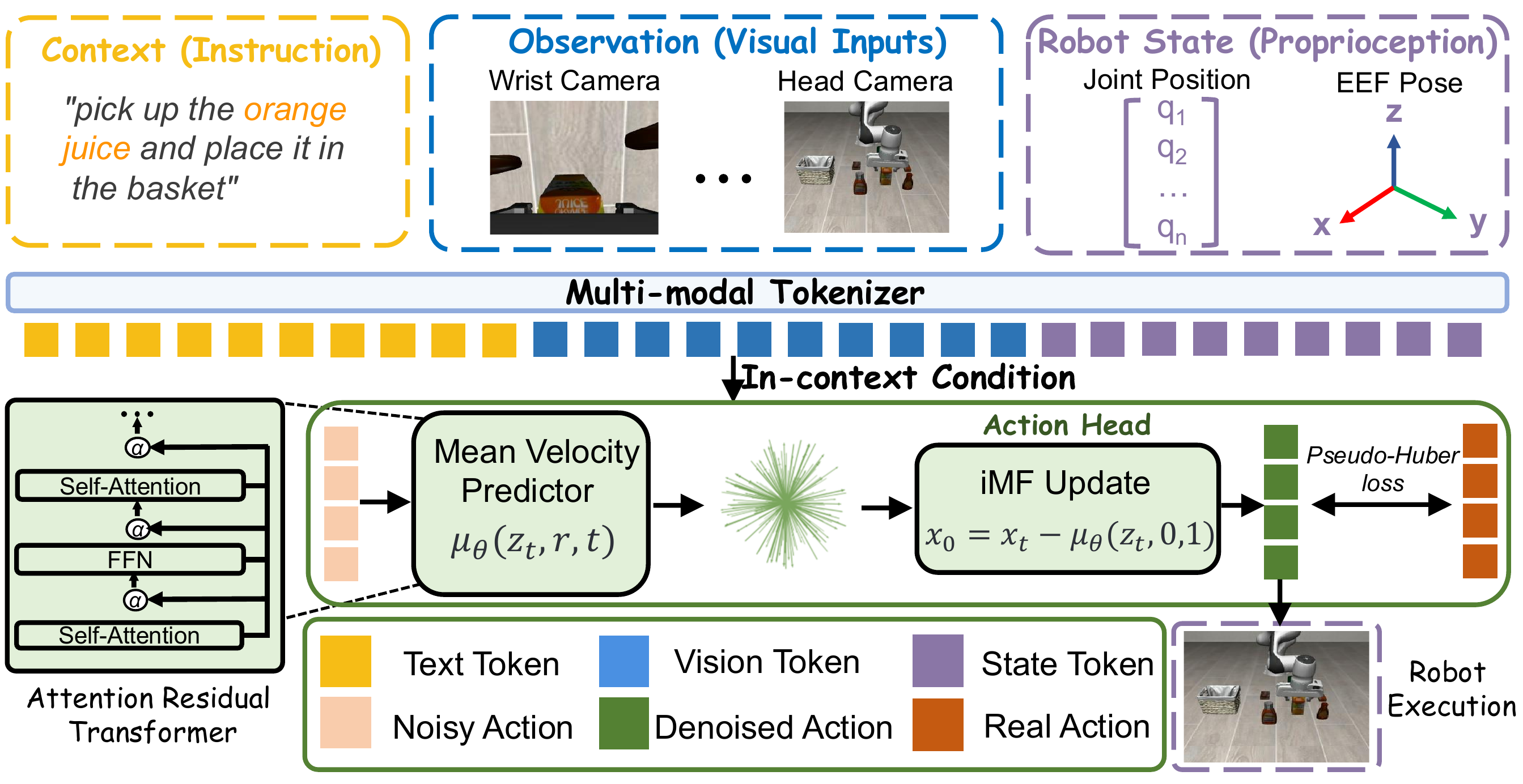}
    \caption{Overview of the \texttt{ReactVLA} framework. Multimodal observations are processed by an Attention Residual Transformer backbone, while the proposed improved Mean Flow (iMF) action head directly transports noisy actions toward the target action distribution through one-to-few-step generation.}
    \label{fig:overview}
\end{figure*}
To reduce the computational burden of iterative sampling, we adapt improved Mean Flow (iMF) to robot action generation. Unlike standard flow matching methods that rely on instantaneous local velocity fields~\cite{lipman2023flow,liu2022flow}, Mean Flow models finite-interval average transport, naturally supporting stable one-to-few-step generation~\cite{geng2025mean,geng2025improved}. Since low-step generation increases the representational demand on each policy evaluation, \texttt{ReactVLA} further introduces Attention Residuals (AttnRes), a dynamic depth-wise routing mechanism for multimodal policy transformers. Instead of uniformly accumulating layer outputs through fixed residual paths~\cite{he2016deep,vaswani2017attention}, AttnRes selectively retrieves informative historical representations across transformer depth~\cite{kimi2026attention}.


By combining iMF action generation with AttnRes-based feature routing, \texttt{ReactVLA} forms a fast and lightweight VLA framework for reactive robot manipulation. As illustrated in Fig.~\ref{fig:lightweight}, our method achieves a favorable trade-off by combining strong success rates with a lightweight model footprint and low inference latency, making it well suited for high-frequency, reactive closed-loop robot control. Our contributions are summarized as follows:
\begin{enumerate}[]
    \item  We propose \texttt{ReactVLA}, a low-latency Vision-Language-Action framework that reformulates robot action generation through improved Mean Flow, enabling stable one-to-few-step generative manipulation.
    
    \item  We introduce an Attention Residual transformer backbone that performs dynamic depth-wise feature routing, mitigating multimodal representation dilution in deep policy transformers under aggressive low-step inference.

    \item  We demonstrate strong manipulation performance across diverse simulation and real-world benchmarks while substantially reducing inference latency. \texttt{ReactVLA} achieves an average 15.2$\times$ higher inference throughput than diffusion-based policies, while outperforming competitive VLA baselines such as SmolVLA and $\pi_0$ on LIBERO and RoboIMI benchmarks.
\end{enumerate}

\section{Related Work}
\label{sec:related_work}

\subsection{Diffusion and Flow Policies for Robot Action Generation}

Generative action models have become a central paradigm for imitation-based robot manipulation. 
Diffusion Policy formulates visuomotor control as conditional denoising over action sequences, enabling expressive multimodal action distributions for manipulation tasks~\cite{chi2023diffusion,ho2020denoising}. 
Subsequent work has scaled this idea with Transformer backbones and VLA models, extending generative policy learning to broader task distributions and language-conditioned settings~\cite{peebles2023scalable,chen2025dreamarrangement,black2024pi0}. 
Despite their effectiveness, these approaches typically rely on iterative sampling, making inference cost a key bottleneck for high-frequency robot control. Flow-based generative policies provide an alternative view by learning continuous vector fields that transport noise to data~\cite{lipman2023flow}. 
Rectified Flow further improves sampling efficiency by encouraging straighter transport paths, reducing the number of function evaluations required at inference~\cite{liu2022flow}. 
These methods motivate fast action generation, but standard flow formulations still model local instantaneous velocities and can suffer from truncation error under extremely low-step sampling. Our \texttt{ReactVLA} follows this fast-generation direction but shifts the action generator from local velocity prediction to finite-interval mean transport. 
By adapting iMF to VLA action generation, our method directly predicts average velocity fields for one-to-few-step action synthesis, targeting the latency requirements of reactive robot manipulation while retaining the expressiveness of generative policies.

\subsection{Residual Connections and Attention Residuals}

Residual connections are a fundamental component of deep neural architectures, enabling stable optimization and gradient propagation across depth~\cite{he2016deep}. 
In large Transformer models, however, standard residual accumulation can progressively dilute intermediate representations due to repeated unweighted feature aggregation~\cite{xiong2020layer}. 
To improve information propagation, prior work has explored gated residual pathways~\cite{srivastava2015highway}, multi-stream residual architectures~\cite{zhu2025hyper,xie2025mhc}, and learned depth-wise aggregation strategies~\cite{pagliardini2024denseformer}. Recent work on Attention Residuals reformulates residual accumulation as an input-dependent retrieval process across transformer depth~\cite{kimi2026attention}. Rather than uniformly summing historical layer outputs, AttnRes enables each layer to dynamically retrieve informative intermediate representations through depth-wise attention routing. While originally developed for large-scale language modeling, its potential for multimodal robot policy learning remains largely unexplored. In particular, low-step generative policies place substantially greater representational demands on each policy evaluation, since action quality must be achieved with only a few generation steps. We therefore introduce AttnRes into a VLA policy and investigate its effectiveness for multimodal manipulation. Our results show that dynamic depth-wise routing helps preserve useful visual, language, and proprioceptive representations throughout the transformer, providing a stronger context representation for low-latency action generation.

\section{Problem Setting and Framework Overview}
\label{sec:policy_formulation}

We consider language-conditioned robot manipulation under a VLA-based imitation learning framework. At each control step $t$, the policy receives a multimodal observation $o_t=\{I_t,s_t,l\}$, where $I_t$, $s_t$, and $l$ denote visual observations, robot proprioceptive states, and the task instruction, respectively. Conditioned on $o_t$, the policy predicts a future action chunk $a_{t:t+H}\sim\pi_\theta(\cdot\mid o_t)$ over horizon $H$. Our goal is to learn a generative policy that produces high-fidelity action chunks with a small inference budget $N_{\text{infer}}$, enabling high-frequency reactive control from streaming sensory feedback.

To reduce the deployment cost of iterative diffusion-style sampling, \texttt{ReactVLA} replaces local denoising with low-step mean flow action generation. As illustrated in Fig.~\ref{fig:overview}, multimodal observations are encoded by an Attention Residual Transformer with dynamic depth-wise feature routing. The resulting representations condition an improved Mean Flow action head, which predicts finite-interval transport from noisy actions to the target action distribution and enables one-to-few-step action generation. Training uses a robust Pseudo-Huber objective for stable low-step action optimization. The key components are described next.

\section{Method}
\label{sec:method}

\subsection{Improved Mean Flow for Action Generation}
\label{sec:imf_action_generation}
\begin{figure}[t]
  \centering
  \includegraphics[width=0.5\textwidth]{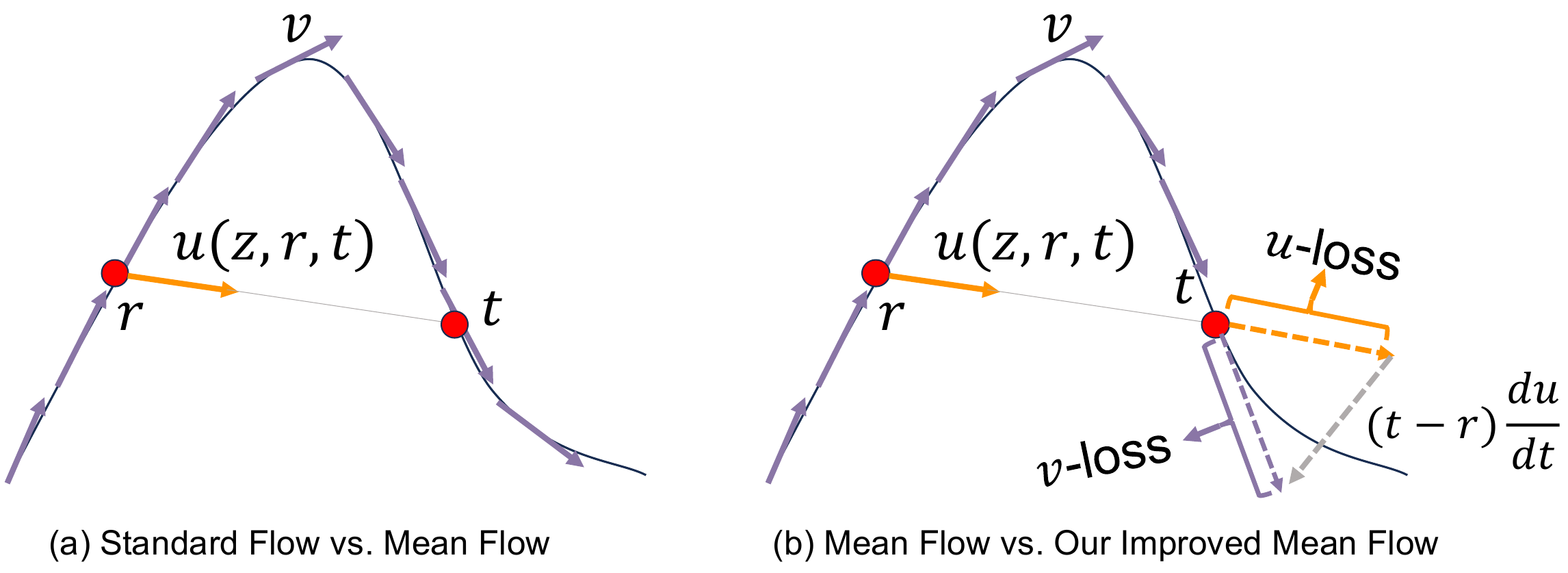}
  \caption{\textbf{Comparison of the Flow Matching, Mean Flow, and Improved Mean Flow.} Standard flow matching models local instantaneous transport, requiring fine-grained integration along curved trajectories. Mean Flow instead learns finite-interval average transport, enabling large-step generation. Improved Mean Flow further stabilizes training through a Jacobian-Vector Product (JVP) correction with stop-gradient constraints.}
  \label{fig:2Meanflow}
\end{figure}

As illustrated in Fig.~\ref{fig:2Meanflow}(a), classical Flow Matching formulates generation through instantaneous velocity fields that locally transport samples along continuous trajectories. While expressive, such local transport requires fine-grained numerical integration and becomes increasingly unstable under aggressive low-step inference, where discretization errors rapidly accumulate along potentially curved trajectories. To alleviate this limitation, Mean Flow~\cite{geng2025mean} replaces instantaneous transport with finite-interval average transport over a macroscopic interval $[r,t]$, where $0 \le r < t \le 1$. Instead of modeling the local velocity field $v(z_t,t)$, Mean Flow directly predicts the average transport velocity 
\begin{equation} 
u(z_t,r,t) = \frac{1}{t-r} \int_r^t v(z_\tau,\tau)d\tau
\label{eq:mean_flow} 
\end{equation} 
where $z_t$ denotes the noisy latent action state. This formulation enables large-step transport across the generation trajectory and substantially reduces the number of inference evaluations required during action generation.

However, as shown in Fig.~\ref{fig:2Meanflow}(b), directly optimizing the mean velocity field can introduce path inconsistency between the conditional transport target and the marginal velocity dynamics, leading to unstable optimization under low-step training. To mathematically capture and resolve this inconsistency, we analyze the relationship between the average velocity field $u(z_t, r, t)$ and the instantaneous velocity field $v(z_t, t)$, where the latent action trajectory $z_t$ at continuous time $t \in [0, 1]$ satisfies the ordinary differential equation (ODE) $\frac{d z_t}{dt} = v(z_t, t)$.

Multiplying both sides of Eq.~\ref{eq:mean_flow} by $(t-r)$ yields the accumulated flow:
\begin{equation}
    (t-r)u(z_t, r, t) = \int_r^t v(z_\tau, \tau) d\tau
\end{equation}
Differentiating this relation with respect to $t$ on both sides using the Leibniz rule yields:
\begin{equation}
    u(z_t, r, t) + (t-r)\frac{d}{dt}u(z_t, r, t) = v(z_t, t)
\end{equation}
Rearranging this equation, we express the average velocity field as:
\begin{equation}
    u(z_t, r, t) = v(z_t, t) - (t-r)\frac{d}{dt}u(z_t, r, t)
    \label{eq:differentiation_identity}
\end{equation}
The total derivative of the average velocity $u(z_t, r, t)$ with respect to time $t$ along the trajectory must account for both the explicit dependence on $t$ and the implicit dependence via the state $z_t$. Applying the chain rule, we write this total derivative as:
\begin{equation}
    \frac{d}{dt}u(z_t, r, t) = \frac{\partial u}{\partial z_t} \frac{dz_t}{dt} + \frac{\partial u}{\partial t} = \frac{\partial u}{\partial z_t} v(z_t, t) + \frac{\partial u}{\partial t}
\end{equation}
This operation can be cast as a Jacobian-Vector Product (JVP) of the average velocity function $u$ with respect to the input variables $(z_t, r, t)$ along the tangent direction $(v(z_t, t), 0, 1)$:
\begin{equation}
    \frac{d}{dt}u(z_t, r, t) = \mathrm{JVP}\big(u; (v(z_t, t), 0, 1)\big)
    \label{eq:jvp_derivative}
\end{equation}
Substituting Eq.~\ref{eq:jvp_derivative} back into Eq.~\ref{eq:differentiation_identity} yields:
\begin{equation}
    u(z_t, r, t) = v(z_t, t) - (t-r)\mathrm{JVP}\big(u; (v, 0, 1)\big)
\end{equation}
In practice, since we train a neural network $u_\theta \approx u$ to represent the average flow, directly optimizing this non-linear relation is highly unstable during training. To address this issue, we adopt Improved Mean Flow (iMF)~\cite{geng2025improved}, which introduces a Jacobian-Vector Product correction to align the average transport prediction with the underlying instantaneous velocity field:
\begin{equation} 
V_\theta(z_t,r,t) = u_\theta(z_t,r,t) + (t-r) \mathrm{JVP}_{\mathrm{sg}} \big( u_\theta; v_\theta \big)
\label{eq:imf} 
\end{equation} 
where $u_\theta(z_t,r,t)$ denotes the predicted average transport velocity and $v_\theta(z_t, t) = u_\theta(z_t, t, t)$ represents the instantaneous velocity prediction obtained from the same network. The stop-gradient operator ($\mathrm{sg}$) stabilizes the nonlinear correction term while preserving consistency between the conditional transport target and the marginal vector field. During deployment, the policy directly performs finite-interval action transport using the learned average-flow predictor $u_\theta$, enabling stable one-to-few-step action generation for low-latency robot control.

\subsection{Attention Residual Policy Transformer}
\label{sec:attnres_policy}
The Attention Residual policy transformer serves as the policy core, taking multimodal observation features and processing them via dynamic depth-wise routing. Specifically, at each control step $t$, the input sequence $X \in \mathbb{R}^{L \times d_{\text{model}}}$ is constructed by concatenating temporal, conditioning, and action tokens:
\begin{equation}
X = \left[\phi(r), \phi(t), W_c C, W_a z_t\right]
\end{equation}
where $d_{\text{model}} = 768$ is the hidden dimension of the transformer, $r, t \in [0,1]$ are the flow time steps mapped to $d_{\text{model}}$-dimensional sinusoidal embeddings $\phi(r), \phi(t) \in \mathbb{R}^{1 \times 768}$ followed by linear projection, $z_t \in \mathbb{R}^{H \times d_a}$ is the noisy action trajectory ($H=16$, $d_a=7$) projected to $W_a z_t \in \mathbb{R}^{16 \times 768}$, and $C$ represents the multimodal conditioning sequence. 

The conditioning sequence $C$ is formed by concatenating visual, textual, and proprioceptive tokens:
\begin{equation}
C = \left[E_{\text{vis}}(I_{\text{agent}}), E_{\text{vis}}(I_{\text{wrist}}), E_{\text{text}}(l), E_{\text{prop}}(s_t)\right]
\end{equation}
where $E_{\text{vis}}(I_{\text{agent}}) \in \mathbb{R}^{N_{\text{vis}} \times d_{\text{vis}}}$ and $E_{\text{vis}}(I_{\text{wrist}}) \in \mathbb{R}^{N_{\text{vis}} \times d_{\text{vis}}}$ represent the spatial features extracted from the agent camera and wrist camera views using a frozen SigLip2 image encoder ($N_{\text{vis}} = 256$, $d_{\text{vis}} = 1152$), $E_{\text{text}}(l) \in \mathbb{R}^{N_{\text{text}} \times d_{\text{text}}}$ represents the textual instruction tokens from a SmolVLM text encoder ($d_{\text{text}} = 2048$), and $E_{\text{prop}}(s_t) \in \mathbb{R}^{1 \times 768}$ is a single proprioceptive token. The proprioceptive token is obtained by linearly projecting the 8-dimensional proprioceptive joint state $s_t \in \mathbb{R}^8$ using a learnable projection matrix $W_{\text{prop}} \in \mathbb{R}^{8 \times 768}$ plus a bias term:
\begin{equation}
E_{\text{prop}}(s_t) = \text{LinearProjection}(s_t) = W_{\text{prop}} s_t + b_{\text{prop}}.
\end{equation}
All conditioning tokens in $C$ are projected to the model dimension $d_{\text{model}} = 768$ via a projection matrix $W_c$ before sequence injection. The resulting total sequence length of the transformer is $L = 2 + 2 N_{\text{vis}} + N_{\text{text}} + 1 + H$.

Deep transformer policies commonly employ residual connections to stabilize optimization across depth. 
However, standard PreNorm residual accumulation uniformly aggregates historical layer outputs, which can progressively dilute multimodal representations in deep policy transformers~\cite{he2016deep,xiong2020layer}. 
A detailed derivation of the PreNorm accumulation formulation is provided in Appendix~\ref{sec:appendix_prenorm}. Such representation dilution becomes particularly problematic under low-step action generation, where each policy evaluation must preserve sufficient visual and proprioceptive information for accurate action prediction. 
To mitigate this issue, our \texttt{ReactVLA} adopts Attention Residuals~\cite{kimi2026attention}, which replaces fixed residual accumulation with dynamic depth-wise feature routing:
\begin{equation}
h_l
=
\sum_{i=0}^{l-1}
\alpha_{i \rightarrow l} v_i,
\qquad
\sum_{i=0}^{l-1}
\alpha_{i \rightarrow l}=1
\end{equation}
where $v_i$ denotes the historical representation from layer $i$, and $\alpha_{i \rightarrow l}$ represents the routing weight assigned by layer $l$.

The routing weights are computed through a lightweight layer-specific query:
\begin{equation}
e_{i \rightarrow l}
=
w_l^\top
\mathrm{RMSNorm}(v_i),
\qquad
\alpha_{i \rightarrow l}
=
\frac{\exp(e_{i \rightarrow l})}
{\sum_{j=0}^{l-1}\exp(e_{j \rightarrow l})}
\end{equation}
where $w_l$ is a learnable routing vector. 
This formulation enables the policy transformer to dynamically preserve informative multimodal features across depth, improving representation stability for low-latency action generation.

\begin{algorithm}[t]
\caption{One training iteration of ReactVLA}
\label{alg:imf_attnres}
\begin{algorithmic}[1]
\Statex \textbf{Input:} action $a$, context $c$, parameters $\theta$, Pseudo-Huber scale $\delta$.
\Statex \textbf{Output:} training loss $\mathcal{L}$
\State $h_{\mathrm{ctx}} \gets \mathrm{AttnResTransformer}_{\theta}(c)$
\State Sample time steps $r,t$ with $r\leq t$, and sample $\epsilon\sim\mathcal{N}(0,I)$
\State $z_t \gets (1-t)a+t\epsilon$, \quad $v_g \gets \epsilon-a$
\State $v_{\theta} \gets u_{\theta}(z_t,t,t\mid h_{\mathrm{ctx}})$
\State $u,\dot{u} \gets \mathrm{JVP}\!\left(u_{\theta}(\cdot\mid h_{\mathrm{ctx}}),(z_t,r,t),(v_{\theta},0,1)\right)$
\State $V_{\theta} \gets u+(t-r)\mathrm{sg}(\dot{u})$
\State $\mathcal{L}\gets \mathrm{PseudoHuber}_{\delta}(V_{\theta}-v_g)$
\State \Return $\mathcal{L}$
\end{algorithmic}
\end{algorithm}


\subsection{Pseudo-Huber Loss for Stable Action Optimization}
\label{sec:pseudo_huber_loss}

Low-step action generation can occasionally produce large transport errors during early training, particularly when the predicted mean flow dynamics are still poorly aligned with the target action trajectory. 
To improve optimization stability, we train the action generator using the Pseudo-Huber loss instead of the standard Mean Squared Error (MSE) objective.

Let
\[
e
=
V_\theta(z_t,r,t)-v_g
\]
denote the prediction error between the corrected mean flow prediction $V_\theta$ and the target transport velocity $v_g$. 
For each error dimension $e_d$, the Pseudo-Huber loss is defined as
\begin{equation}
\rho_\delta(e_d)
=
\delta^2
\left(
\sqrt{
1+
\left(
\frac{e_d}{\delta}
\right)^2
}
-1
\right)
\end{equation}
where $\delta>0$ controls the transition between quadratic and linear error penalization. 
The overall training objective is computed as
\begin{equation}
\mathcal{L}(\theta)
=
\frac{1}{D}
\sum_{d=1}^{D}
\rho_\delta(e_d)
\end{equation}

To understand why the Pseudo-Huber loss is crucial for stabilizing \texttt{ReactVLA}'s training, we analyze its gradient propagation behavior. Under the training objective, the parameters $\theta$ are updated via the gradient:
\begin{equation}
\frac{\partial \mathcal{L}(\theta)}{\partial \theta} = \frac{1}{D} \sum_{d=1}^{D} \frac{\partial \rho_\delta(e_d)}{\partial e_d} \frac{\partial e_d}{\partial \theta}
\end{equation}
where the derivative of the Pseudo-Huber loss with respect to the error dimension $e_d$ is given by:
\begin{equation}
\frac{\partial \rho_\delta(e_d)}{\partial e_d} = \frac{e_d}{\sqrt{1 + \left( \frac{e_d}{\delta} \right)^2}}
\label{eq:huber_gradient}
\end{equation}
Equation~\ref{eq:huber_gradient} reveals that as the error magnitude $|e_d| \to \infty$, the term $\frac{\partial \rho_\delta(e_d)}{\partial e_d} \to \pm \delta$, meaning the gradient factor is strictly bounded within $[-\delta, \delta]$. In contrast, standard Mean Squared Error (MSE) loss yields $\frac{\partial \mathcal{L}_{\mathrm{MSE}}}{\partial e_d} = e_d$, which scales linearly without bound.

This bounding behavior is critical when optimizing policies with JVP corrections. The corrected velocity prediction $V_\theta$ (Eq.~\ref{eq:imf}) contains the first-order spatial derivatives of $u_\theta$ to form the JVP. Consequently, backpropagating the loss to compute the parameter gradient $\frac{\partial e_d}{\partial \theta}$ requires taking derivatives through the JVP term, which introduces second-order derivatives of the neural network $u_\theta$. During the early phases of training, the predicted mean flow velocity fields are unaligned and highly non-linear, which easily produces extremely large prediction errors $e_d$ and high-variance second-order gradients in $\frac{\partial e_d}{\partial \theta}$. Under an MSE loss, these two factors multiply, causing catastrophic gradient spikes that destabilize training (as observed in Fig.~\ref{fig:loss_comparison}). By bounding the error feedback via Eq.~\ref{eq:huber_gradient}, the Pseudo-Huber loss acts as an inherent gradient-clipping mechanism at the loss level, preventing large transport deviations from propagating high-magnitude updates through the second-order JVP path, thereby ensuring smooth and stable optimization.

A detailed analysis of the asymptotic behavior of the Pseudo-Huber formulation is provided in Appendix~\ref{sec:appendix_huber}. Algorithm~\ref{alg:imf_attnres} further summarizes one training iteration of \texttt{ReactVLA}, including multimodal feature routing, improved Mean Flow prediction, JVP correction, and robust action optimization. The exact time-step sampling strategy is described in Appendix~\ref{sec:Generation}.

\begin{table*}[h]
\centering
\small
\setlength{\tabcolsep}{8pt}
\caption{Simulation results on LIBERO benchmark. Success rates (\%) for various policies across different task categories (Spatial, Object, Goal, Long-horizon). We include the inference latency (ms) for direct comparison.}
\label{tab:libero_results}
\begin{tabular}{llcccccr}
\toprule
\multirow{2}{*}{Method} & \multirow{2}{*}{\# Params} & \multicolumn{5}{c}{Success Rate (\%) — Simulation} & \multirow{2}{*}{Latency (ms)} \\
\cmidrule(r){3-7}
& & Spatial & Object & Goal & Long & Avg. & \\
\midrule
Diffusion Policy~\cite{chi2023diffusion} & 0.46B & 78.3 & 92.5 & 68.3 & 50.5 & 72.4 & 178.82 \\
OpenVLA~\cite{kim2025openvla} & 6.74B & 84.7 & 88.4 & 79.2 & 53.7 & 76.5 & 115.0 \\
$\pi_0$ (Paligemma-3B)~\cite{black2024pi0} & 5.97B & 87.0 & 63.0 & 89.0 & 48.0 & 71.8 & 94.3 \\
$\pi_0$ ~\cite{black2024pi0} & 4.03B & 90.0 & 86.0 & 95.0 & 73.0 & 86.0 & 93.4 \\
SmolVLA~\cite{shukor2025smolvla} & \textbf{0.24B} & 87.0 & 93.0 & 88.0 & 63.0 & 82.8 & 71.1 \\
SmolVLA~\cite{shukor2025smolvla} & 0.45B & 90.0 & \textbf{96.0} & \textbf{92.0} & 71.0 & 87.3 & 74.1 \\
\midrule
\textbf{ReactVLA (Ours)} & 0.39B & \textbf{93.0} & 95.0 & \textbf{92.0} & \textbf{72.0} & \textbf{88.0} & \textbf{18.3} \\
\bottomrule
\end{tabular}
\end{table*}

\begin{table*}[t]
\centering
\small
\caption{Performance comparison on RoboIMI simulation tasks (Peg-in-Socket Insertion and Object Transfer) over 100 evaluation rollouts. We report the average reward for each task and the policy inference latency (ms).}
\label{tab:roboimi_results}
\begin{tabular}{lccc}
\toprule
Method & Peg-in-Socket Reward & Object Transfer Reward & Latency (ms) \\
\midrule
ACT~\cite{zhao2023learning} & 289.60 & 115.64 & \textbf{4.5} \\
SmolVLA~\cite{shukor2025smolvla} & -- & -- & -- \\
Diffusion Policy~\cite{chi2023diffusion} & 1019.39 & 319.20 & 398.0 \\
\textbf{ReactVLA (Ours)} & \textbf{1513.56} & \textbf{526.22} & 15.1 \\
\bottomrule
\end{tabular}
\end{table*}

\section{Experiments}
\label{sec:experiments}

\subsection{Setup and Metrics}
\label{sec:setup_metrics}

To validate the efficiency and precision of our proposed \texttt{ReactVLA} framework, we perform extensive evaluations across two simulation benchmarks and one real-world robotic platform:
\begin{itemize}[]
    \item \textbf{LIBERO Benchmark:} A standard suite for VLA robot learning models. To ensure a fair comparison, the training and evaluation protocols (including the evaluation metrics and episode setups) strictly follow the pipeline of SmolVLA~\cite{shukor2025smolvla}.
    \item \textbf{RoboIMI Simulation Platform:} A custom-built, dual-arm robotic simulation environment designed for coordinated, tight-tolerance manipulation. We evaluate all comparative policies across 100 evaluation trials per task and report the average cumulative reward and inference latency. We evaluate on two challenging tasks:
    \begin{itemize}
        \item \textit{Peg-in-Socket:} One arm picks up a wooden peg from the table and inserts it into a rectangular socket held dynamically by the other robotic arm.
        \item \textit{Object Transfer:} One arm picks up a wooden block from the table and transfers it to the gripper of the other arm.
    \end{itemize}
    \item \textbf{Diana Robot Platform:} A physical single-arm robotic platform equipped with a Diana 7 robotic arm from Agile Robots. We evaluate our framework and baseline methods over 20 runs per task on two real-world manipulation tasks: Orange Pick-and-Place and Block Stack. The robot is required to grasp an orange and place it stably on a plate in Orange Pick-and-Place, and to stack a wooden block on a pre-existing block stack in Block Stack. We report the task success rate and physical inference latency.
\end{itemize}

\subsection{Simulation Experiments on LIBERO}
\label{sec:libero_results}

We evaluate our proposed \texttt{ReactVLA} on the widely used LIBERO benchmark~\cite{shukor2025smolvla}, following the multi-task protocols of SmolVLA~\cite{shukor2025smolvla}. As shown in Table~\ref{tab:libero_results}, our policy achieves a strong average success rate of $88.0\%$, outperforming representative VLA models across most task categories.
Crucially, while standard diffusion- or flow-based VLA models require multiple sampling or integration steps to generate smooth actions, \texttt{ReactVLA} achieves this using only $2$ steps. This reduction lowers the policy latency to $18.3$ ms ($54.6$ Hz) while maintaining strong manipulation performance, demonstrating the effectiveness of low-step action generation for efficient robot control.

\subsection{Simulation Experiments on RoboIMI}
\label{sec:roboimi_results}

We evaluate our model on the custom-built RoboIMI simulation platform, which serves as a bimanual robotic environment to test coordinated and high-precision manipulation. We perform rollouts on two tasks: Peg-in-Socket and Object Transfer. The overall performance comparison of different policy models across both tasks is summarized in Table~\ref{tab:roboimi_results}. The results show that our proposed \texttt{ReactVLA} significantly outperforms all baseline methods in terms of average reward, while executing at exceptionally low inference latencies (approximately $15$ ms). On the challenging peg-in-socket task, our method achieves an average reward of 1513.56, outperforming Diffusion Policy by $48.5\%$ while reducing latency from $398.0$ ms to $15.1$ ms. On the bimanual handover task, it achieves an average reward of 526.22, exceeding the standard Diffusion Policy baseline at $319.20$ with a low latency of $15$ ms. Notably, the SmolVLA baseline failed to achieve successful policy convergence on this dual-arm setup, consistently yielding zero rewards due to training complexities; its performance metrics are therefore omitted and denoted as ``--'' in the table. These results demonstrate that \texttt{ReactVLA} successfully resolves the speed-quality trade-off, enabling both highly accurate and real-time reactive control for coordinated bimanual manipulation.

\subsection{Ablation Studies}
\label{sec:ablation_studies}


\begin{figure}
    \centering
      \includegraphics[width=0.6\linewidth]{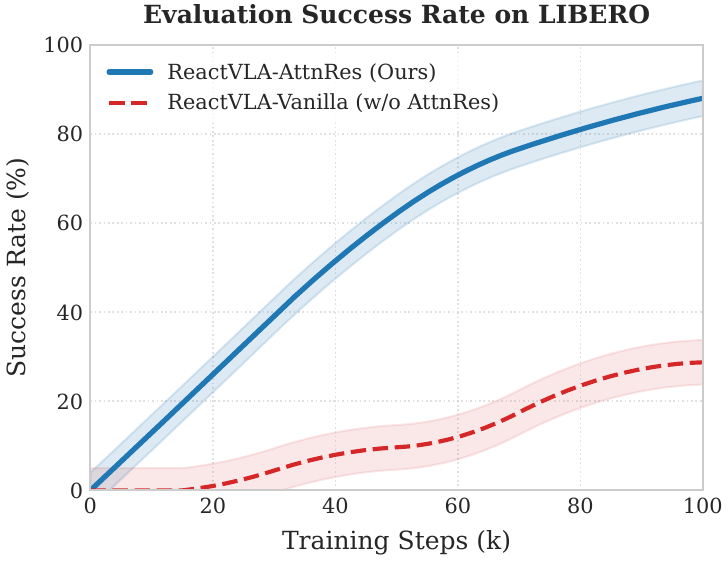}
      \caption{\textbf{Ablation Study on Attention Residuals in LIBERO Simulation}. We compare \texttt{ReactVLA} against the vanilla baseline that uses standard uniform additive residual connections.}
      \label{fig:3ablation1}
      \vspace{-0.6\baselineskip}
\end{figure}

\begin{figure}[t]
    \centering
    \includegraphics[width=0.5\textwidth]{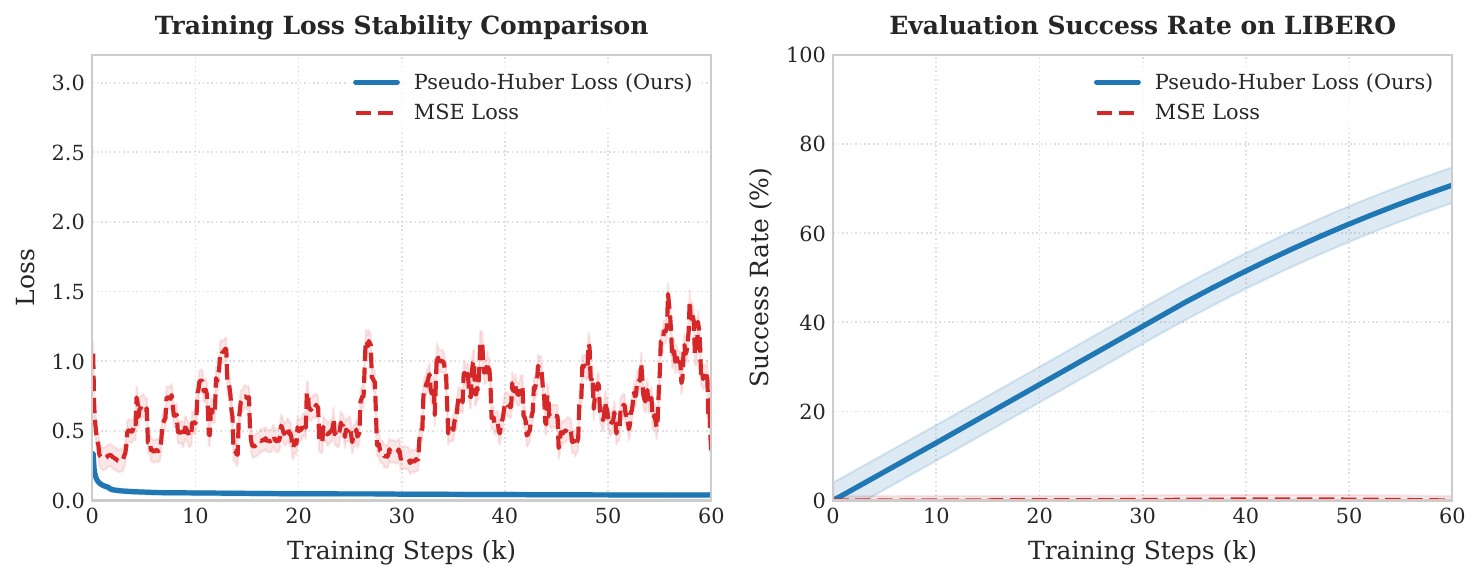}
    \caption{\textbf{Ablation Study on Loss Formulations.} Comparison of training loss trajectories (left) and evaluation success rates (right) between Pseudo-Huber Loss (Ours) and MSE Loss over 60k training steps.   
    }
    \label{fig:loss_comparison}
\end{figure}

\begin{figure*}[t]
  \centering
    \includegraphics[width=\textwidth]{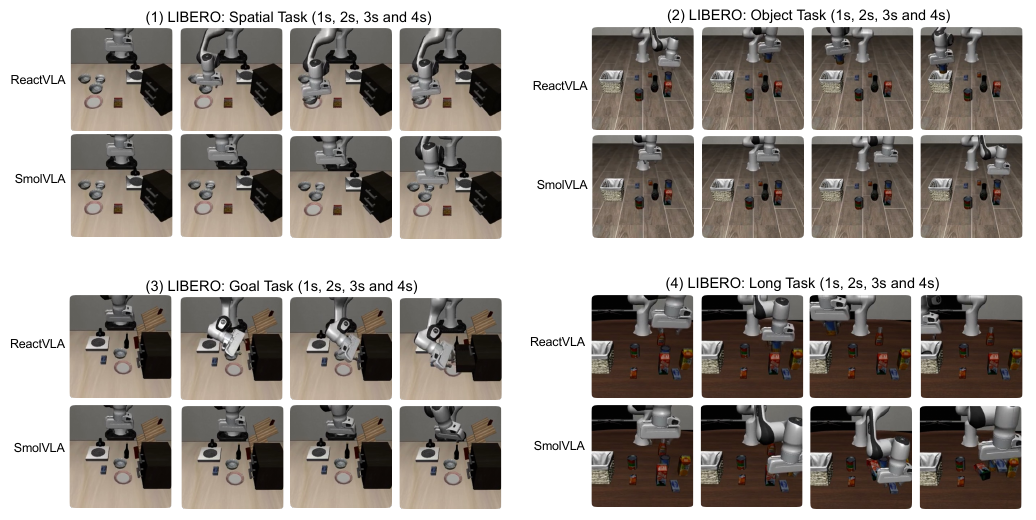}

  \caption{\textbf{Qualitative Comparison of Manipulation Rollouts.} In simulation, \texttt{ReactVLA} generates smoother trajectories compared to SmolVLA. The rollout keyframes are sampled at 1s, 2s, 3s, and 4s due to the relatively short overall duration of LIBERO simulation trials.}
  \label{fig:main_visualizations}
\end{figure*}

\begin{figure*}[t]
    \includegraphics[width=\textwidth]{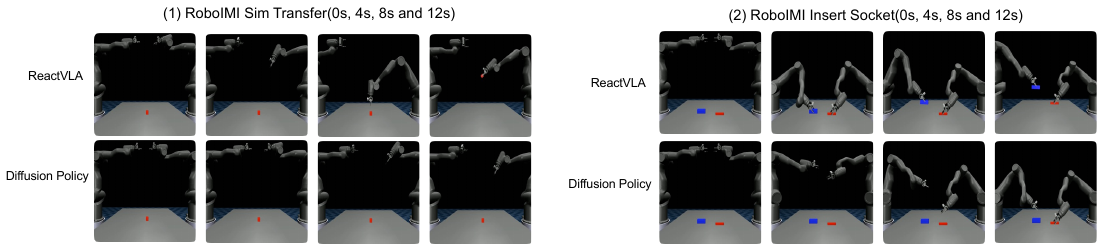}

  \caption{\textbf{Qualitative Rollouts in RoboIMI Dual-Arm Tasks.} \texttt{ReactVLA} coordinates both robotic arms smoothly during peg-in-socket insertion and block handover compared to the standard Diffusion Policy. The keyframes are chosen at 0s, 4s, 8s, and 12s because RoboIMI simulation episodes have a much longer execution duration, typically lasting dozens of seconds.}
  \label{fig:main_visualizations_roboimi}
\end{figure*}
\textbf{Attention Residuals:} To analyze the impact of the Attention Residual structure on the model's training dynamics, we compare our full model, \texttt{ReactVLA}, which incorporates dynamic layer-specific softmax residuals, against a vanilla baseline variant, \texttt{ReactVLA-Vanilla} (where AttnRes is replaced with standard uniform additive residuals). As shown in Fig.~\ref{fig:3ablation1}, \texttt{ReactVLA} achieves significantly improved final performance. The evaluation success rate of \texttt{ReactVLA} (right panel) rises much faster, surpassing $50\%$ within the first $40$k steps and converging to a peak success rate of $88.0\%$, whereas the vanilla model converges more slowly to a final success rate of only $28.7\%$. This confirms that dynamic depth-wise attention helps prevent feature dilution and eases optimization.

    
    

\textbf{Loss Formulations (Pseudo-Huber vs. MSE):} We further investigate the influence of action-generation loss formulations by comparing the Pseudo-Huber loss used in our framework against the standard Mean Squared Error (MSE) loss. As illustrated in Fig.~\ref{fig:loss_comparison}, the choice of loss function plays a critical role in training stability. The model trained with MSE loss suffers from frequent loss spikes and severe gradient instability, which is primarily induced by explosive Jacobian-Vector Products  during backpropagation when handling large action deviations. In contrast, the model trained with Pseudo-Huber loss exhibits a highly stable and smooth convergence curve. By sub-linearly scaling large errors, the Pseudo-Huber loss effectively bounds the vector magnitudes within the JVP computations, providing robust gradient behavior during action trajectory optimization and preventing training destabilization.

\subsection{Qualitative Evaluation and Visualization}
\label{sec:qualitative}

To qualitatively evaluate the performance of \texttt{ReactVLA}, we visualize and compare the rollout trajectories of our policy against baseline models in simulation. Fig.~\ref{fig:main_visualizations} illustrates the qualitative rollout frames on the LIBERO benchmark tasks, comparing our method against SmolVLA. Fig.~\ref{fig:main_visualizations_roboimi} shows the rollout comparison on the RoboIMI simulation platform tasks against Diffusion Policy. Crucially, because \texttt{ReactVLA} achieves a much faster inference speed and lower latency, it can execute more policy control steps within the same physical time duration. Consequently, at the same physical elapsed time, \texttt{ReactVLA} progresses faster and is significantly closer to completing the task, whereas the baseline models lag behind due to their slower inference. For instance, in the RoboIMI bimanual block handover task (Fig.~\ref{fig:main_visualizations_roboimi}), at $t = 8$\,s, \texttt{ReactVLA} has already coordinated both arms to secure the block and initiate the handover process, whereas the standard Diffusion Policy is still slow to react and remains in the initial approach stage. Similarly, for the LIBERO benchmark tasks (Fig.~\ref{fig:main_visualizations}), because of the rapid execution enabled by our low-latency design, \texttt{ReactVLA} completes the manipulation sequence significantly ahead of SmolVLA within the same physical time (visualized at keyframes 1s, 2s, 3s, and 4s).

\begin{figure}[t]
    \centering
    \includegraphics[width=0.5\textwidth]{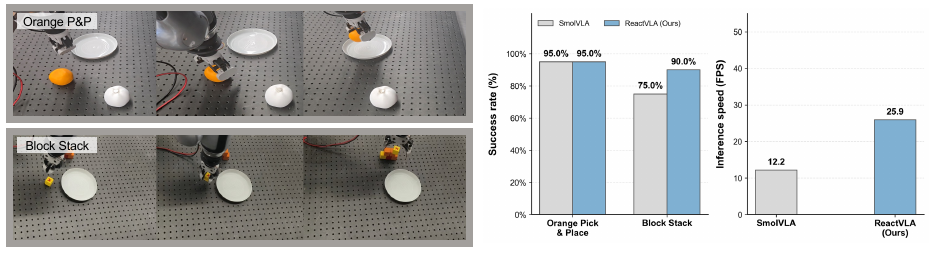}
    \caption{\textbf{Real-World Robot Experiments.} (Left) Demonstration of the Orange Pick-and-Place and Block Stack tasks on the Diana 7 robotic arm. (Right) Performance comparison (success rate vs. latency) of our method against baseline models in real-world deployment.}
    \label{fig:real_world_exp}
\end{figure}
\subsection{Real-World Robot Experiments }
\label{sec:real_world_results}

To verify the transferability and low-latency performance of \texttt{ReactVLA} on physical systems, we deploy our framework on the Diana 7 robotic arm for both the Orange Pick-and-Place and Block Stack tasks, as illustrated in Fig.~\ref{fig:real_world_exp}. These tasks require precise end-effector control and responsive visuomotor feedback. We collect 50 teleoperated demonstrations for each task to perform model training. The inputs to \texttt{ReactVLA} include egocentric wrist camera views, a third-person camera feed, and the robot’s joint states, while the output is target joint commands executed at 60 Hz. In physical testing, our low-step policy achieves an average latency of 38.6 ms, allowing the robot to adapt reactively to perturbations. Over 20 evaluation trials on the Orange Pick-and-Place task, both \texttt{ReactVLA} and the SmolVLA baseline achieve a high success rate of 95\% (19/20 runs). However, on the more challenging Block Stack task, \texttt{ReactVLA} achieves a success rate of 90\% (18/20 runs), outperforming SmolVLA,  which drops to 75\% (15/20 runs) due to error accumulation in its slower control loop. Furthermore, SmolVLA exhibits a higher inference latency of 82.2 ms. This demonstrates that our fast action generation is critical for enabling tighter, real-time closed-loop feedback and more fluent reactive control in physical environments.

\section{Conclusion}
\label{sec:conclusion}

We presented \texttt{ReactVLA}, a low-latency, lightweight VLA framework for reactive robot manipulation. 
Our approach combines an improved Mean Flow action generator with an Attention Residual transformer backbone, enabling one-to-few-step action generation while preserving the multimodal representations required for accurate policy execution. 
Across LIBERO, RoboIMI, and real-world robot experiments, \texttt{ReactVLA} achieves strong manipulation performance with substantially reduced inference latency, demonstrating that efficient generative action modeling can be realized without sacrificing control quality. By shifting from local iterative denoising to finite-interval transport prediction and dynamic depth-wise feature routing, \texttt{ReactVLA} offers a practical path toward real-time, closed-loop robot manipulation. Despite promising results, several limitations remain. 
First, our real-world evaluation focuses on a controlled tabletop manipulation setting with a relatively small number of demonstrations. Extending \texttt{ReactVLA} to more diverse environments, longer-horizon tasks, and highly cluttered scenes remains an important direction for future work. 
Second, although all methods were evaluated on identical hardware within each benchmark to ensure fair comparisons, different benchmark suites were conducted on different GPU platforms. Consequently, absolute latency measurements should be interpreted within each benchmark, while the reported relative speedups more directly reflect the computational efficiency of the underlying action-generation algorithms.




\bibliographystyle{IEEEtran}
\bibliography{bibtex/bib/reference}

\section{Appendix}

\subsection{PreNorm Representation Accumulation}
\label{sec:appendix_prenorm}

In a standard Transformer architecture with Pre-Layer Normalization (PreNorm), the hidden state $h_l \in \mathbb{R}^d$ entering the $l$-th layer is updated recurrently. Let $f_{l-1}$ denote the transformation computed by the $(l-1)$-th layer. The standard PreNorm update can be written as:
\begin{equation}
h_l = h_{l-1} + f_{l-1}(h_{l-1}).
\label{eq:appendix_prenorm_step}
\end{equation}
Unrolling this recurrence from the initial layer input $h_1$, which represents the observation and instruction context embeddings, gives:
\begin{align}
h_2 &= h_1 + f_1(h_1), \\
h_3 &= h_2 + f_2(h_2) = h_1 + f_1(h_1) + f_2(h_2), \\
&\ \vdots \nonumber \\
h_l &= h_1 + \sum_{i=1}^{l-1} f_i(h_i).
\label{eq:appendix_prenorm_sum}
\end{align}
This formulation reveals a structural limitation of standard PreNorm residual connections: all preceding layer outputs are accumulated with fixed unit weights. As the depth increases, the norm of the accumulated hidden state can grow with depth, making the incremental contribution of each newly added layer relatively smaller compared with the already accumulated representation. Consequently, deeper layers may have limited ability to introduce sufficiently distinct or task-relevant transformations, effectively reducing the useful depth of the network and leading to representation dilution. The Attention Residual mechanism addresses this limitation by replacing fixed residual accumulation with softmax-normalized, input-dependent routing over preceding layer representations, enabling each layer to selectively aggregate the most relevant historical features for final action decoding.

\subsection{Asymptotic Limits of the Pseudo-Huber Loss}
\label{sec:appendix_huber}

The Pseudo-Huber loss is defined for each dimension $d \in \{1, \dots, D\}$ of the velocity prediction error $e_d \in \mathbb{R}$ as:
\begin{equation}
    \rho_\delta(e_d) = \delta^2 \left( \sqrt{1 + \left(\frac{e_d}{\delta}\right)^2} - 1 \right)
    \label{eq:appendix_pseudo_huber_def}
\end{equation}
where $\delta > 0$ is the scale parameter. We analyze the asymptotic behavior of $\rho_\delta(e_d)$ under the two extremes: small errors and large errors.

\paragraph{Case 1: Small Errors ($|e_d| \ll \delta$)}
When the error magnitude is much smaller than the scale parameter $\delta$, the ratio $\frac{e_d}{\delta}$ is small. We can apply the Taylor series expansion for the function $\sqrt{1 + x}$ around $x = 0$, which is:
\begin{equation}
    \sqrt{1 + x} = 1 + \frac{1}{2}x - \frac{1}{8}x^2 + \mathcal{O}(x^3).
\end{equation}
Setting $x = \left(\frac{e_d}{\delta}\right)^2$, we have:
\begin{equation}
\begin{aligned}
\sqrt{1 + \left(\frac{e_d}{\delta}\right)^2}
&=
1
+ \frac{1}{2}\left(\frac{e_d}{\delta}\right)^2
- \frac{1}{8}\left(\frac{e_d}{\delta}\right)^4  \\
&\quad
+ \mathcal{O}\left(
\left(\frac{e_d}{\delta}\right)^6
\right).
\end{aligned}
\end{equation}
Substituting this expansion back into \ref{eq:appendix_pseudo_huber_def}:
\begin{align}
    \rho_\delta(e_d) &= \delta^2 \left( 1 + \frac{1}{2}\left(\frac{e_d}{\delta}\right)^2 - \frac{1}{8}\left(\frac{e_d}{\delta}\right)^4 + \mathcal{O}\left(\left(\frac{e_d}{\delta}\right)^6\right) - 1 \right) \nonumber \\
    &= \frac{1}{2} e_d^2 - \frac{1}{8\delta^2} e_d^4 + \mathcal{O}\left(\frac{e_d^6}{\delta^4}\right)
\end{align}
Thus, to the first-order approximation:
\begin{equation}
    \rho_\delta(e_d) \approx \frac{1}{2} e_d^2 \quad \text{for } |e_d| \ll \delta
\end{equation}
which is equivalent to the standard L2 (Mean Squared Error) loss scaled by $1/2$.

\paragraph{Case 2: Large Errors ($|e_d| \gg \delta$)}
When the error magnitude is much larger than $\delta$, the ratio $\frac{e_d}{\delta}$ is very large. We can approximate the square root term as:
\begin{equation}
\begin{aligned}
\sqrt{1 + \left(\frac{e_d}{\delta}\right)^2}
&=
\sqrt{
\left(\frac{e_d}{\delta}\right)^2
\left(
1 + \left(\frac{\delta}{e_d}\right)^2
\right)
}
\\
&=
\frac{|e_d|}{\delta}
\sqrt{
1 + \left(\frac{\delta}{e_d}\right)^2
}
\end{aligned}
\end{equation}
Since $\frac{\delta}{|e_d|} \ll 1$, we can apply the Taylor expansion $\sqrt{1+y} \approx 1 + \frac{1}{2}y$ with $y = \left(\frac{\delta}{e_d}\right)^2$:
\begin{equation}
    \sqrt{1 + \left(\frac{\delta}{e_d}\right)^2} \approx 1 + \frac{\delta^2}{2e_d^2}
\end{equation}
Thus, we obtain:
\begin{equation}
    \sqrt{1 + \left(\frac{e_d}{\delta}\right)^2} \approx \frac{|e_d|}{\delta} \left( 1 + \frac{\delta^2}{2e_d^2} \right) = \frac{|e_d|}{\delta} + \frac{\delta}{2|e_d|}
\end{equation}
Substituting this back into \ref{eq:appendix_pseudo_huber_def} yields:
\begin{align}
    \rho_\delta(e_d) &\approx \delta^2 \left( \frac{|e_d|}{\delta} + \frac{\delta}{2|e_d|} - 1 \right) \nonumber \\
    &= \delta |e_d| - \delta^2 + \frac{\delta^3}{2|e_d|}
\end{align}
As $|e_d| \to \infty$, the term $\frac{\delta^3}{2|e_d|}$ approaches zero, so the dominant terms are:
\begin{equation}
    \rho_\delta(e_d) \approx \delta |e_d| - \delta^2 \quad \text{for } |e_d| \gg \delta
\end{equation}
which represents a linear L1-like loss with slope $\delta$ and a constant offset. This formulation prevents outlier errors from generating excessively large gradients.

\subsection{Details of Model Architecture}
\label{sec:appx_training}

We instantiate \texttt{ReactVLA} as a vision-language-conditioned action flow model operating directly within the continuous action space. Each multimodal observation step incorporates dual-view RGB inputs, a global third-person view (\texttt{agentview\_image}) and an eye-in-hand view (\texttt{robot0\_eye\_in\_hand\_image}), both rendered at a $256\times256$ resolution, alongside an 8-dimensional robot proprioceptive joint state and a natural language task instruction. The policy predicts 7-dimensional relative-control actions over a temporal trajectory horizon of $H=16$ and executes an action chunk size of $K=8$ steps per inference cycle, conditioned on a single historical observation step ($n_{\mathrm{obs}}=1$).

For visual-language conditioning, we utilize a decoupled multimodal embedding pipeline. Specifically, we employ a frozen \texttt{SigLip2} architecture as the visual encoder to extract spatial features from the input images, which are resized and normalized appropriately. Concurrently, the natural language task instruction is processed by the text transformer layers of the \texttt{SmolVLM} model, serving as our dedicated language encoder to capture semantic context. The visual tokens from both camera views, the textual tokens from the language encoder, and a single proprioceptive state token (obtained by linearly projecting the 8-dimensional joint state to match the transformer's hidden dimension) are concatenated to form a unified multimodal conditioning sequence $C$. These conditioning tokens are linearly mapped to the action head hidden dimension of 768 prior to sequence injection.

The action generation head consists of a 16-block Attention Residual (AttnRes) Transformer operating on a combined sequence comprising temporal embeddings, the conditioning context tokens, and 16 action latent tokens:
\begin{equation}
X = \left[\phi(r), \phi(t), W_c C, W_a z\right]
\end{equation}
where $r$ and $t$ represent the source and target flow time steps, $\phi(\cdot)$ denotes a sinusoidal time embedding followed by a linear projection, $C$ is the multimodal conditioning sequence, and $z\in\mathbb{R}^{16\times7}$ corresponds to the current raw action trajectory latent. 

The AttnRes backbone utilizes a hidden size of 768 with 8 query heads and 8 key-value heads, executing standard multi-head attention (MHA). Positional formatting relies on Rotary Position Embeddings (RoPE) with a base frequency of $10{,}000$, supplemented by Root Mean Square Normalization (RMSNorm, $\epsilon=10^{-6}$) and a dropout rate of 0.05 across both the embedding and attention layers. Each AttnRes block is composed of a RoPE-augmented self-attention sublayer and a SwiGLU feed-forward network (FFN) sublayer with an intermediate hidden dimension of 2048.

In contrast to standard residual architectures that sequentially accumulate hidden states uniformly across layers, the AttnRes mechanism maintains an indexed cache of all preceding layer representations. Prior to entering the $m$-th attention or feed-forward sublayer, a token-wise depth attention operation routes information across historical branches:
\begin{equation}
\begin{aligned}
\bar h^{(m)}_{\tau}
&=
\sum_{j<m}
\alpha^{(m)}_{j,\tau} h^{(j)}_{\tau},
\\
\alpha^{(m)}_{j,\tau}
&=
\operatorname{softmax}_{j}
\left(
q_m^\top
\operatorname{RMSNorm}
\left(h^{(j)}_{\tau}\right)
\right).
\end{aligned}
\end{equation}
where $q_m$ represents a learned structural pseudo-query specific to the $m$-th sublayer, and $\tau$ denotes the token index. The sublayer output is subsequently appended as a newly indexed hidden branch, and the final representation exiting the backbone is obtained via a summation of all accumulated branches. Following the transformer blocks, the last 16 action tokens are isolated, passed through an RMSNorm layer, and projected back into the 7-dimensional relative control space.

\subsection{Implementation Details of Action Generation}
\label{sec:Generation}

Action generation operates entirely within the native, normalized trajectory space. Given a normalized raw action trajectory $x \in \mathbb{R}^{H\times d_a}$ with $H=16$ and $d_a=7$, the network directly learns and samples continuous action trajectories without domain transformations.

During the training phase, we sample Gaussian noise $e\sim\mathcal{N}(0,I)$ alongside two distinct flow time steps $r,t\in[0,1]$ drawn from a logit-normal distribution ($\mu = -0.4, \sigma = 1.0$). The sampled intervals are sorted such that $0\le r\le t\le1$, with a 0.5 probability of setting $r=t$ to incorporate standard flow-matching targets. The interpolated noisy trajectory latent $z_t$ is formulated via linear interpolation:
\begin{equation}
z_t = (1-t)x + t e .
\end{equation}
The network $f_\theta(z_t,r,t,C)$ parameterizes the velocity field conditioned on the multimodal context prefix $C$. To stabilize backpropagation through complex flow paths, \texttt{ReactVLA} implements an improved Mean Flow (iMF) objective augmented with a Jacobian-Vector Product (JVP) correction term:
\begin{equation}
\hat v = u_\theta + (t-r) \mathcal{D}_{z} f_\theta(z_t,r,t,C) \left[ f_\theta(z_t,t,t,C), 0, 1 \right]^\top
\end{equation}
where $u_\theta=f_\theta(z_t,r,t,C)$ represents the base predicted velocity, and $\mathcal{D}_{z}$ denotes the Jacobian operator with respect to the latent trajectory state. The fundamental optimization target is defined as $e-x$. To mitigate gradient spikes caused by demonstration noise, training relies directly on the Pseudo-Huber loss formulation applied to the raw trajectory dimensions with a threshold parameter $\delta=1$:
\begin{equation}
\rho_\delta(\epsilon) = \delta^2 \left(\sqrt{1+(\epsilon/\delta)^2}-1\right).
\end{equation}
Auxiliary regularizations, such as padding masks, frequency reweighting, and semigroup consistency constraints, are omitted in our configuration.

At inference time, action generation initiates from random Gaussian noise $z_1\sim\mathcal{N}(0,I)$ sampled within the normalized action domain. We deploy a highly efficient two-step inference regime over the discrete time grid $1.0 \rightarrow 0.5 \rightarrow 0.0$. For each integration interval $(t_i,r_i)$, the latent state is sequentially updated via an Euler-style flow integration step:
\begin{equation}
z_{r_i} = z_{t_i} - (t_i-r_i) f_\theta(z_{t_i}, r_i, t_i, C).
\end{equation}
Upon reaching the final time boundary, the optimized latent vector $z_0$ directly represents the predicted action trajectory in the normalized action space. Given an observation horizon of $n_{\mathrm{obs}}=1$ and a targeted execution chunk of $K=8$, the policy slices and emits the first eight actions from the decoded 16-step trajectory ($z_{0, 0:7}$). In a closed-loop deployment cycle, these actions are queued and executed sequentially at the environment's control rate; a fresh multimodal action chunk is generated immediately upon queue depletion. Lastly, an unnormalization post-processor scales the generated relative commands back to the native operational boundaries of the robotic environment.


\subsection{Evaluation of Simulation Experiments}
\label{sec:A20}
\subsubsection{Implementation details of Simulation tasks}

\textbf{LIBERO multi-task setup.} We evaluate our method on the standard LIBERO benchmark, specifically utilizing four multi-task suites: LIBERO-Spatial, LIBERO-Object, LIBERO-Goal, and LIBERO-Long, each containing 10 distinct manipulation tasks. To thoroughly evaluate the multi-task scaling capability of our model, we aggregate the demonstrations from all four suites into a single large-scale unified dataset comprising 40 distinct manipulation tasks, upon which the policy is trained jointly. The policies utilize dual-view RGB observations from the agent-view camera and the eye-in-hand camera, together with the robot's proprioceptive joint state and the task language embedding. For our proposed \texttt{ReactVLA}, we train entirely on this official aggregated multi-task demonstration set. Unlike standard diffusion- or flow-based VLA models that require multiple expensive denoising or sampling steps, our framework predicts actions using only 2 inference steps, significantly accelerating simulation rollouts while maintaining a strong average success rate.

\textbf{RoboIMI setup and tasks.} We evaluate our framework on the custom-built RoboIMI simulation platform, which serves as a dual-arm robotic environment specifically designed for coordinated, tight-tolerance manipulation. We conduct evaluations across two challenging tasks: 
(1) \textit{Peg-in-Socket}, where one robotic arm picks up a wooden peg from a randomized position on the tabletop and precisely inserts it into a rectangular socket held dynamically by the opposing arm; 
(2) \textit{Object Transfer}, where one arm picks up a wooden block from a randomized position on the table and executes a coordinated handover to the gripper of the other arm. 
The policies take multimodal inputs, including RGB observations from three cameras, a third-person global camera and two wrist-mounted cameras (one on each arm), along with the complete dual-arm proprioceptive joint states and gripper execution states. All comparative models, including ACT, Diffusion Policy, and \texttt{ReactVLA}, are trained on the demonstrations generated by an automated script policy for each task. To thoroughly verify the real-time coordination and precision under tight tolerances, we evaluate each policy across 100 randomized simulation trials per task, reporting the average cumulative reward and inference latency directly from the log files.

\textbf{Reward Mechanism and Formulations.} The reward computation for both tasks is grounded in the physics engine's contact dynamics, implemented by querying the MuJoCo contact array \texttt{mj\_data.contact} at each simulation timestep $t$. Interacting geometric shapes (\textit{geoms}) are extracted to construct contact pairs, which are then evaluated against task-specific rules to determine the per-step reward $r_t$. Crucially, the final performance metric reported in our evaluation is the episodic cumulative reward, defined as the time-aggregated sum over the entire rollout horizon $T$:
\begin{equation}
R = \sum_{t=1}^{T} r_t .
\end{equation}
Under this temporal accumulation formulation, once the robot successfully achieves and sustains a specific sub-goal or milestone state, the corresponding step reward $r_t$ is persistently added to the cumulative total $R$ at every subsequent timestep until the episode terminates. Consequently, early task completion coupled with stable state maintenance results in a continuous accumulation of the high-tier step rewards, allowing the cumulative metric $R$ to effectively encapsulate both the temporal efficiency and operational stability of different policies. The two environments adopt fundamentally distinct scoring philosophies and underlying pair-matching implementations to compute the step reward $r_t$:
\paragraph{Staged reward in \textit{Object Transfer}}
This task simulates a sequential, single-object state machine that tracks the
transfer of a \texttt{red\_box} from the right arm to the left arm. It uses a
stepwise overwriting reward with a maximum per-step score of $r_t=4$:
\begin{equation}
\small
r_t =
\begin{cases}
1, & \text{right grasps box without invalid table contact}, \\
2, & \text{right lifts box airborne}, \\
3, & \text{left contacts box}, \\
4, & \text{left secures box airborne}.
\end{cases}
\end{equation}
Technically, contact pairs are represented as ordered tuples, making the
evaluation sensitive to the MuJoCo geometric ID ordering, e.g.,
\texttt{(geomA, geomB)} versus \texttt{(geomB, geomA)}.

\paragraph{Additive reward in \textit{Peg-in-Socket}}
This task requires asynchronous dual-arm coordination to manipulate two separate
objects, a \texttt{red\_peg} and a \texttt{blue\_socket}, for mid-air assembly. It
uses a multi-objective additive reward with a maximum per-step score of $r_t=5$:
\begin{equation}
\small
\begin{aligned}
r_t
=&\ \mathbb{I}(\text{left contacts socket})
 + \mathbb{I}(\text{right contacts peg}) \\
&+ \mathbb{I}(\text{socket airborne})
 + \mathbb{I}(\text{peg airborne}) \\
&+ \mathbb{I}(\text{peg contacts internal socket pin}),
\end{aligned}
\end{equation}
where $\mathbb{I}(\cdot)$ denotes the indicator function.

This design dichotomy reflects the underlying task structures: \textit{Object Transfer} uses a strict staged progression to guide a linear object hand-off pipeline, whereas \textit{Peg-in-Socket} leverages an additive structure to independently incentivize decoupled sub-goals before enforcing the final tight-tolerance alignment.


\subsection{Evaluation of Real-world Experiments}
\label{sec:A30}
\subsubsection{Implementation details of Real-world tasks}
\label{sec:A31}

\textbf{Robot platform.} We conduct all real-world experiments on a Diana 7 single-arm robotic arm from Agile Robots. The policy outputs joint position commands and a gripper command at each control step. The proprioceptive input contains the robot's joint states and the gripper execution state. To evaluate the robustness and spatial generalization of the policies, we introduce small perturbations to the robotic arm's initial poses within a localized range before commencing each trial, and vary the target orange positions across different runs, while keeping the remaining workspace layout constant.

\textbf{Orange P\&P.} Orange Pick-and-Place (Orange P\&P) is a representative tabletop manipulation task in which the robot must grasp an orange from the tabletop and place it stably onto a target plate. We collect 50 teleoperated demonstrations on the Diana 7 platform for model training. The final deployed policies utilize two camera views: an agent-view RGB camera and a wrist-mounted RGB camera. The agent view provides global semantic context of the tabletop scene and the target plate, while the wrist camera provides close-range visual feedback to facilitate reliable gripper alignment during grasping. The visual inputs are resized and processed before being passed to the visual encoder. The models are trained entirely on the 50 expert demonstrations without any additional real-world rollout data.

\textbf{Block Stack.} Block Stack is a high-precision tabletop stacking task where a small stack of wooden blocks is already present on the tabletop. The robotic arm must grasp another block from the tabletop and place it stably on top of the pre-existing stack without causing the stack to collapse. We collect 50 teleoperated demonstrations on the Diana 7 platform for model training. The policy utilizes the same dual-camera setup (agent-view and wrist camera) and observations as the Orange P\&P task. The model is also trained entirely on these 50 demonstrations, requiring high spatial accuracy and closed-loop responsiveness to achieve stable stacking.

\textbf{Real-world deployment protocol.} During real-world deployment, both \texttt{ReactVLA} and the SmolVLA baseline process the multimodal observation streams online and generate target joint commands executed at a high control frequency of 60 Hz. Specifically, the number of inference steps for \texttt{ReactVLA} is set to 5 during physical testing to guarantee a sufficiently smooth and stable action trajectory. All evaluated methods follow the identical initial reset protocol and strict task success criteria. A trial of the Orange P\&P task is quantified as successful if and only if the orange is securely grasped, transferred, and placed stably on the target plate without any object drops or unintended collisions.



\end{document}